\pdfoutput=1

\documentclass[11pt]{article}
\usepackage{tikz}

\usepackage[]{acl}
\usepackage{amsmath}
\usepackage{times}
\usepackage{latexsym}
\usepackage{multirow}
\usepackage{fancybox}

\usepackage{paralist}
\usepackage[T1]{fontenc}

\usepackage[utf8]{inputenc}

\usepackage{microtype}

\usepackage{pgf}
\usepackage{tikz}
\usepackage{hhline}
\usepackage{collcell}

\def\colorModel{hsb} 

\newcommand\ColCell[1]{
  \pgfmathparse{#1<50?1:0}  
    \ifnum\pgfmathresult=0\relax\color{white}\fi
  \pgfmathsetmacro\compA{0}      
  \pgfmathsetmacro\compB{#1/100} 
  \pgfmathsetmacro\compC{1}      
  \edef\x{\noexpand\centering\noexpand\cellcolor[\colorModel]{\compA,\compB,\compC}}\x #1
  } 
\newcolumntype{E}{>{\collectcell\ColCell}c<{\endcollectcell}}  

\newcommand*{\MinNumber}{0}%
\newcommand*{\MaxNumber}{1}

\definecolor{col2}{RGB}{255, 201, 166}
\definecolor{col1}{RGB}{255, 243, 217}

\newcommand{\ApplyGradients}[1]{%
        \pgfmathsetmacro{\PercentColor}{100.0*(#1-\MinNumber)/(\MaxNumber-\MinNumber)}
        \hspace{-0.63em}{
        \begin{tikzpicture}
        \node [fill={col1!\PercentColor!col2}, rounded corners=3pt, baseline=(current bounding box)]{\textcolor{black}{#1}};
        \end{tikzpicture}}
        \vspace{-1pt}
        
}

\newcolumntype{R}{>{\collectcell\ApplyGradients}c<{\endcollectcell}}

\title{Numeric Magnitude Comparison Effects in Large Language Models}

\makeatletter
\def\thanks#1{\protected@xdef\@thanks{\@thanks
        \protect\footnotetext{#1}}}
\makeatother

\newcommand{\gtlogo}{\raisebox{3.4pt}{\includegraphics[scale=0.04]{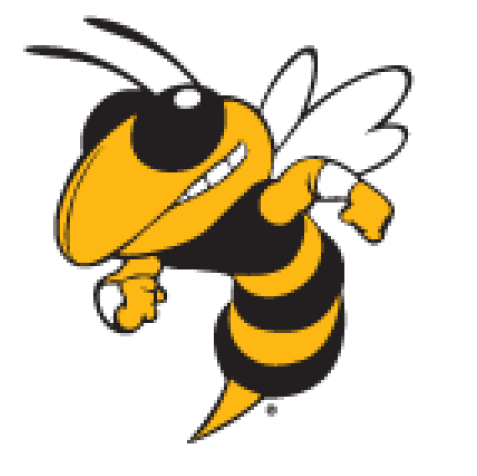}}}

\author{Raj Sanjay Shah, Vijay Marupudi, Reba Koenen,
Khushi Bhardwaj, Sashank Varma
 \\ 
 Georgia Institute of Technology \gtlogo \\
 \textcolor{darkblue}{{\{\href{mailto:rajsanjayshah@gatech.edu}{rajsanjayshah},
\href{mailto:vijaymarupudi@gatech.edu}{vijaymarupudi}, \href{mailto:rkoenen3@gatech.edu}{rkoenen3}, \href{mailto:khushi.bhardwaj@gatech.edu}{khushi.bhardwaj}, \href{mailto:varma@gatech.edu}{varma}\}@gatech.edu}
}}

\newcommand{\raj}[1]{}
\newcommand{\sv}[1]{}
\newcommand{\vijay}[1]{}
\newcommand{\reba}[1]{}
\newcommand{\khushi}[1]{}

\begin{document}
\maketitle
\begin{abstract}
Large Language Models (LLMs) do not differentially represent numbers, which are pervasive in text. In contrast, neuroscience research has identified distinct neural representations for numbers and words. In this work, we investigate how well popular LLMs capture the magnitudes of numbers (e.g., that $4 < 5$) from a behavioral lens.
Prior research on the representational capabilities of LLMs evaluates whether they show human-level performance, for instance, high overall accuracy on standard benchmarks. 
Here, we ask a different question, one inspired by cognitive science: How closely do the number representations of LLMs correspond to those of human language users, who typically demonstrate the \emph{distance}, \emph{size}, and \emph{ratio} effects?
We depend on a linking hypothesis to map the similarities among the model embeddings of number words and digits to human response times. 
The results reveal surprisingly human-like representations across language models of different architectures, despite the absence of the neural circuitry that directly supports these representations in the human brain.
This research shows the utility of understanding LLMs using behavioral benchmarks and points the way to future work on the number representations of LLMs and their cognitive plausibility.

\end{abstract}

\section{Introduction}
\begin{figure*}[htp]
\centering
\includegraphics[trim = {2cm 1cm 2cm 2cm}, width=0.8\textwidth]{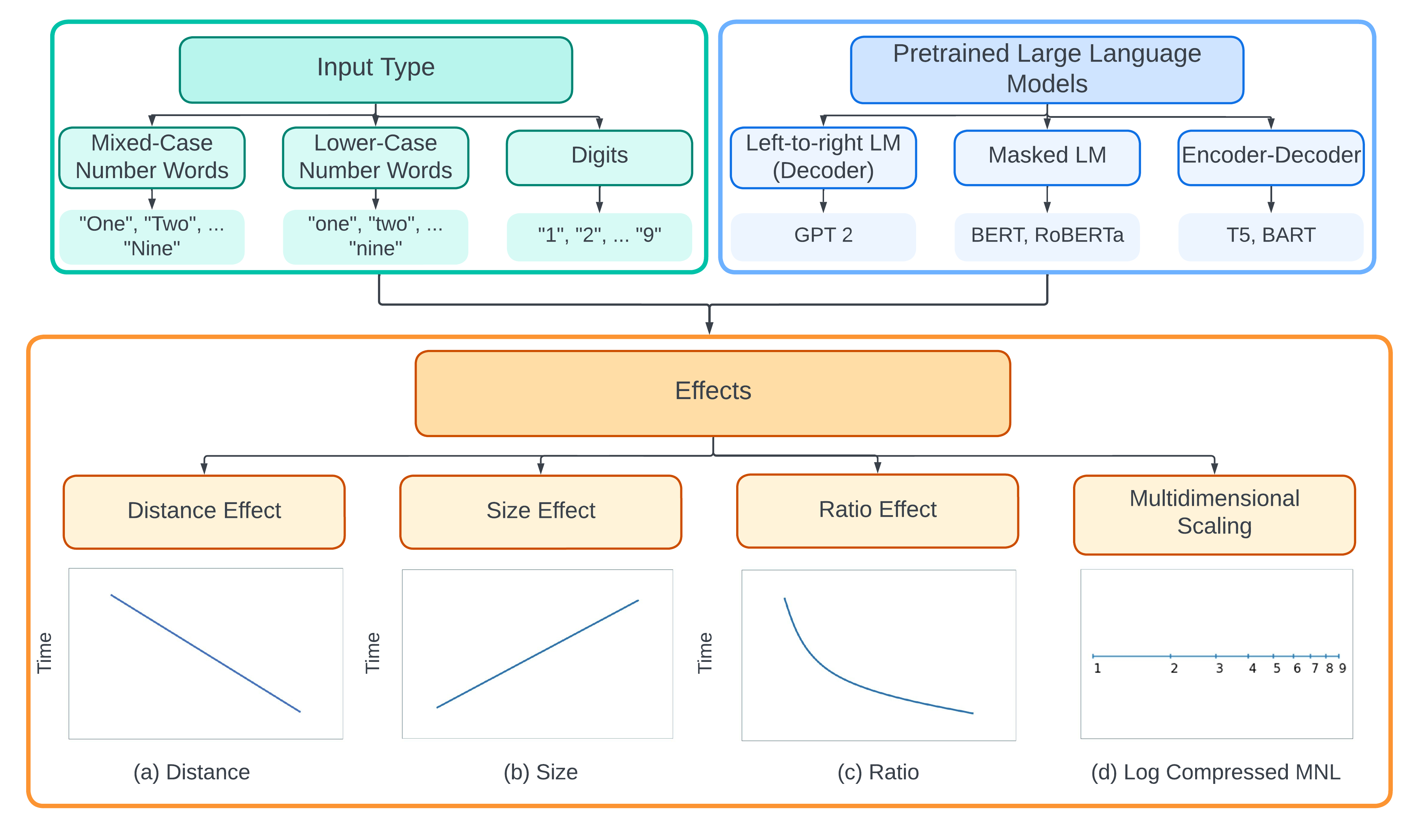}
\caption{The input types, LLMs, and effects in this study. The three effects are depicted in an abstract manner in sub-figures (a), (b), (c).}
\label{fig:introduction0}
\end{figure*}

Humans use symbols – number words such as “three” and digits such as “3” – to quantify the world. How humans understand these symbols has been the subject of cognitive science research for half a century. The dominant theory is that people understand number symbols by mapping them to mental representations, specifically \emph{magnitude representations} \cite{moyerTimeRequiredJudgements1967}. This is true for both number words (e.g., “three”) and digits (e.g., “3”). These magnitude representations are organized as a “mental number line” (MNL), with numbers mapped to points on the line as shown in Figure \ref{fig:introduction0}d. Cognitive science research has revealed that this representation is present in the minds of young children \cite{ansariNeuralCorrelatesSymbolic2005} and even non-human primates \cite{niederCodingCognitiveMagnitude2003}. Most of this research has been conducted with numbers in the range 1-9, in part, because corpus studies have shown that 0 belongs to a different distribution \cite{dehaeneCrosslinguisticRegularitiesFrequency1992} and, in part, because larger numbers require parsing place-value notation \cite{nuerkDecadeBreaksMental2001}, a cognitive process beyond the scope of the current study.

Evidence for this proposal comes from magnitude comparison tasks in which people are asked to compare two numbers (e.g., 3 vs. 7) and judge which one is greater (or lesser). Humans have consistently exhibited three effects that suggest recruitment of magnitude representations to understand numbers: the distance effect, the size effect, and the ratio effect \cite{moyerTimeRequiredJudgements1967, merkleyUsingEyeTracking2010}. We review the experimental evidence for these effects, shown in Figure \ref{fig:introduction0}, in LLMs.
Our \emph{behavioral benchmarking} approach shifts the focus from what abilities LLMs have in an absolute sense to whether they successfully mimic human performance characteristics. This approach can help differentiate between human tendencies captured by models and the model behaviors due to training strategies.
Thus, the current study bridges between Natural Language Processing (NLP), computational linguistics, and cognitive science.

\subsection{Effects of Magnitude Representations}
\label{effects_1}
Physical quantities in the world, such as the brightness of a light or the loudness of a sound, are encoded as logarithmically scaled magnitude representations \cite{fechnerElementsPsychophysics1860}. Research conducted with human participants and non-human species has revealed that they recruit many of the same brain regions, such as the intra-parietal sulcus, to determine the magnitude of symbolic numbers \cite{billockHonorFechnerObey2011, niederRepresentationNumberBrain2009}.

Three primary magnitude representation effects have been found using the numerical comparison task in studies of humans. First, comparisons show a \emph{distance effect}: The greater the distance $|x - y|$ between the numbers $x$ vs. $y$, the faster the comparison \cite{moyerTimeRequiredJudgements1967}. Thus, people compare 1 vs. 9 faster than 1 vs. 2. This is shown in abstract form in Figure \ref{fig:introduction0}a. This effect can be explained by positing that people possess an MNL. When comparing two numbers, they first locate each number on this representation, determine which one is ``to the right'', and choose that number as the greater one. Thus, the farther the distance between the two points, the easier (and thus faster) the judgment. 

Second, comparisons show a \emph{size effect}: Given two comparisons of the same distance (i.e., of the same value for $|x - y|$), the smaller the numbers, the faster the comparison \cite{parkmanTemporalAspectsDigit1971}. For example, 1 vs. 2 and 8 vs. 9 both have the same distance (i.e., $|x - y| = 1$), but the former involves smaller numbers and is therefore the easier (i.e., faster) judgment. The size effect is depicted in abstract form in Figure \ref{fig:introduction0}b. This effect also references the MNL, but a modified version where the points are \emph{logarithmically compressed}, i.e., the distance from 1 to $x$ is proportional to $\log(x)$; see Figure \ref{fig:introduction0}d. To investigate if a logarithmically compressed number line is also present in LLMs, we use multidimensional scaling \cite{mds_Ding2018} on the cosine distances between number embeddings.

Third, comparisons show a \emph{ratio effect}: The time to compare two numbers $x$ vs. $y$ is a decreasing function of the ratio of the larger number over the smaller number, i.e., $\frac{\max(x, y)}{\min(x, y)} $ \cite{halberdaIndividualDifferencesNonverbal2008}. This function is nonlinear, as depicted in abstract form in Figure \ref{fig:introduction0}c. Here, we assume that this function is a negative exponential, though other functional forms have been proposed in the cognitive science literature. The ratio effect can also be explained by the logarithmically compressed MNL depicted in Figure \ref{fig:introduction0}d.

These three effects — distance, size, and ratio — have been replicated numerous times in studies of human adults and children, non-human primates, and many other species \cite{cantlonMathMonkeysDeveloping2012, cohenkadoshAreNumbersSpecial2008}. The MNL model in Figure \ref{fig:introduction0}d accounts for these effects (and many others in the mathematical cognition literature). Here, we use LLMs to evaluate a novel scientific hypothesis: that the MNL representation of the human mind is latent in the statistical structure of the linguistic environment, and thus learnable. Therefore, there is less need to posit pre-programmed neural circuitry to explain magnitude effects.

\subsection{LLMs and Behavioral Benchmarks}
Modern NLP models are pre-trained on large corpora of texts from diverse sources such as Wikipedia~\cite{wiki} and the open book corpus~\cite{bookcorpus}. LLMs like BERT~\cite{bert}, ROBERTA~\cite{roberta} and GPT-2~\cite{gpt2} learn contextual semantic vector representations of words. These models have achieved remarkable success on NLP benchmarks \cite{glue}. They can perform as well as humans on a number of language tests such as semantic verification \cite{bhatiaTransformerNetworksHuman2022} and semantic disambiguation \cite{lakeWordMeaningMinds2021}.

Most benchmarks are designed to measure the absolute performance of LLMs, with higher accuracy signaling “better” models. Human or superhuman performance is marked by exceeding certain thresholds. Here, we ask not whether LLMs can perform well or even exceed human performance at tasks, but whether they show the same \emph{performance characteristics} as humans while accomplishing the same tasks. We call these \emph{behavioral benchmarks}. The notion of behavioral benchmarks requires moving beyond accuracy (e.g., scores) as the dominant measure of LLM performance. 

As a test case, we look at the distance, size, and ratio effects as behavioral benchmarks to determine whether LLMs understand numbers as humans do, using magnitude representations. This requires a \emph{linking hypothesis} to map measures of human performance to indices of model performance. Here, we map human response times on numerical comparison tasks to similarity computations on number word embeddings.

\subsection{Research Questions}
 \label{label:research_questions}
The current study investigates the number representations of LLMs and their alignment with the human MNL. It addresses five research questions: 
\begin{compactenum}
    \item Which LLMs, if any, capture the distance, size, and ratio effects exhibited by humans? 
    \item How do different layers of LLMs vary in exhibiting these effects? 
    \item How do model behaviors change when using larger variants (more parameters) of the same architecture?
    \item Do the models show implicit numeration ("four" = "4"), i.e., do they exhibit these effects equally for all number symbol types or more for some types (e.g., digits) than others (e.g., number words)?
    \item Is the MNL representation depicted in Figure \ref{fig:introduction0}d latent in the representations of the models? 
\end{compactenum}

\section{Related Work}
Research on the numerical abilities of LLMs focuses on several aspects of mathematical reasoning \cite{thawani-etal-2021-representing}, such as magnitude comparison, numeration ~\cite{naik-etal-2019-exploring, wallace_2019}, arithmetic word problems~\cite{awp_1,awp_2}, exact facts ~\cite{exact_facts}, and measurement estimation ~\cite{approx_measurement}. The goal is to improve performance on application-driven tasks that require numerical skills. Research in this area typically attempts to (1) understand the numerical capabilities of pre-trained models and (2) propose new architectures that improve numerical cognition abilities \cite{Geva_at_al, dua-etal-2019-drop}.

Our work also focuses on the first research direction: probing the numerical capabilities of pre-trained models. Prior research by \citet{wallace_2019} judges the numerical reasoning of various contextual and non-contextual models using different tests (e.g., finding the maximum number in a list, finding the sum of two numbers from their word embeddings, decoding the original number from its embedding). These tasks have been presented as evaluation criteria for understanding the numerical capabilities of models. \citet{Numeracy_for_language} change model architectures to treat numbers as distinct from words. Using perplexity score as a proxy for numerical abilities, they argue that this ability reduces model perplexity in neural machine translation tasks. Other work focuses on finding numerical capabilities through building QA benchmarks for performing discrete reasoning \cite{dua-etal-2019-drop}. Most research in this direction casts different tasks as proxies of numerical abilities of NLP systems \cite{downstream_training, dua-etal-2019-drop, Numeracy_for_language, wallace_2019, awp_1, awp_2}.

An alternative approach by \citet{naik-etal-2019-exploring} tests multiple non-contextual task-agnostic embedding generation techniques to identify the failures in models' abilities to capture the magnitude and numeration effects of numbers. Using a systematic foundation in cognitive science research, we build upon their work in two ways: we (1) use contextual embeddings spanning a wide variety of pre-training strategies, and (2) evaluate models by comparing their behavior to humans. Our work looks at numbers in an abstract sense, and is relevant for the grounding problem studied in artificial intelligence and cognitive science \cite{Symbol_grounding}.
\begin{table}[!ht]
    \centering
    \resizebox{0.46\textwidth}{!}{%
    \begin{tabular}{llll}
\hline
        Model & Category & \multicolumn{2}{c}{\begin{tabular}[c]{@{}c@{}}  Size \end{tabular}} \\ 
        & & Base & Large\\
        \hline
        \multicolumn{1}{l}{\begin{tabular}[l]{@{}l@{}}  BERT~\cite{bert} \end{tabular}}
        & Encoder  & 110M & 340M   \\ 
        \multicolumn{1}{l}{\begin{tabular}[l]{@{}l@{}}RoBERTA~\cite{roberta} \end{tabular}}
         & Encoder      & 125M & 355M \\ 
        \multicolumn{1}{l}{\begin{tabular}[l]{@{}l@{}} XLNET~\cite{xlnet} \end{tabular}}
         & Auto-regressive Encoder & 110M & 340M  \\ 
        \multicolumn{1}{l}{\begin{tabular}[l]{@{}l@{}} GPT-2~\cite{gpt2} \end{tabular}}
         & Auto-regressive Decoder & 117M & 345M \\ 
        \multicolumn{1}{l}{\begin{tabular}[l]{@{}l@{}} T5~\cite{t5} \end{tabular}}
          & Encoder & 110M & 335M  \\ 
         \multicolumn{1}{l}{\begin{tabular}[l]{@{}l@{}} BART~\cite{bart}  \end{tabular}}
          & Encoder-Decoder & 140M & 406M\\ 
          
        \hline
    \end{tabular}
    }
    \caption{Popular Language Models }
    \label{tab:models}
\end{table}
\section{Experimental Design}


        
The literature lacks adequate experimental studies demonstrating magnitude representations of numbers in LLMs from a cognitive science perspective.
The current study addresses this gap. We propose a general methodology for mapping human response times to similarities computed over LLM embeddings. We test for the three primary magnitude representation effects described in section \ref{effects_1}.

\subsection{Linking Hypothesis}
In studies with human participants, the distance, size, and ratio effects are measured using reaction time. Each effect depends on the assumption that when comparing which of two numbers $x$ and $y$ is relatively easy, humans are relatively fast, and when it is relatively difficult, they are relatively slow. The ease or difficulty of the comparison is a function of $x$ and $y$: $|x - y|$ for the distance effect, $\min(x, y)$ for the size effect, and $\frac{\max(x, y)}{\min(x, y)}$ for the ratio effect. LLMs do not naturally make reaction time predictions. Thus, we require a \emph{linking hypothesis} to estimate the relative ease or difficulty of comparisons for LLMs. Here we adopt the simple assumption that \emph{the greater the similarity of two number representations in an LLM, the longer it takes to discriminate them, i.e., to judge which  one is greater (or lesser)}. 

We calculate the \emph{similarity} of two numbers based on the similarity of their vector representations. Specifically, the representation of a number for a given layer of a given model is the vector of activation across its units. There are many similarity metrics for vector representations \cite{similarity_measures}: Manhattan, Euclidean, cosine, dot product, etc. Here, we choose a standard metric in distributional semantics: the cosine of the angle between the vectors \cite{richieSimilarityJudgmentCategories2021}. This reasoning connects an index of model function (i.e., the similarity of the vector representations of two numbers) to a human behavioral measure (i.e., reaction time). Thus, the more similar the two representations are, the less discriminable they are from each other, and thus the longer the reaction time to select one over the other. 

\subsection{Materials}
For these experiments, we utilized three formats for number representations in LLMs: lowercase number words, mixed-cased number words (i.e., the first letter is capitalized), and digits. These formats enable us to explore variations in input tokens and understand numeration in models. Below are examples of the three input types:

\begin{compactitem}
    \item "one", "two", "three", "four" ... "nine"
    \item "One", "Two", "Three", "Four" ... "Nine"
    \item "1", "2", "3", "4" ... "9"
\end{compactitem}
As noted in the Introduction, prior studies of the distance, size and ratio effects in humans have largely focused on numbers ranging from 1 to 9. Our input types are not-affected by tokenization methods as the models under consideration have each input as a separate token.

\subsection{Large Language Models - Design Choices}

Modern NLP models are pre-trained on a large amount of unlabeled textual data from a diverse set of sources. This enables LLMs to learn contextually semantic vector representations of words. 
We experiment on these vectors to evaluate how one specific dimension of human knowledge - \emph{number sense} - is captured in different model architectures.

We use popular large language models from Huggingface's Transformers library~\cite{huggingface} to obtain vector representations of numbers in different formats. Following the work by~\citet{popular_models} to determine popular model architectures, we select models from three classes of architectural design: encoder models (e.g., BERT~\cite{bert}), auto-regressive models (e.g., GPT-2~\cite{gpt2}), and encoder-decoder models (e.g., T5~\cite{t5}). The final list of models is provided in Table \ref{tab:models}.

\textbf{Operationalization}:
We investigate the three number magnitude effects as captured in the representations of each layer of the six models for the three number formats. For these experiments, we consider only the obtained hidden layer outputs for the tokens corresponding to the input number word tokens. We ignore the special prefix and suffix tokens of models (e.g., the [cls] token in BERT) for uniformity among different architectures. 
For the T5-base model, we use only the encoder to obtain model embedding.
All models tested use a similar number of model parameters (around 110-140 million parameters). For our studies, we arbitrarily choose the more popular BERT uncased variant as opposed to the cased version. We compare the two models in Appendix section \ref{case_uncased} for a complete analysis, showing similar behaviors in the variants. Model size variations for the same architecture are considered in the Appendix section \ref{variants} to show the impact of model size on the three effects. 

\section{Magnitude Representation Effects in LLMs}
\label{effects}



\subsection{The Distance Effect}

\begin{figure*}[!h]
\centering
\includegraphics[trim = {5.5cm 2cm 0cm 1cm}, width=0.82\textwidth]{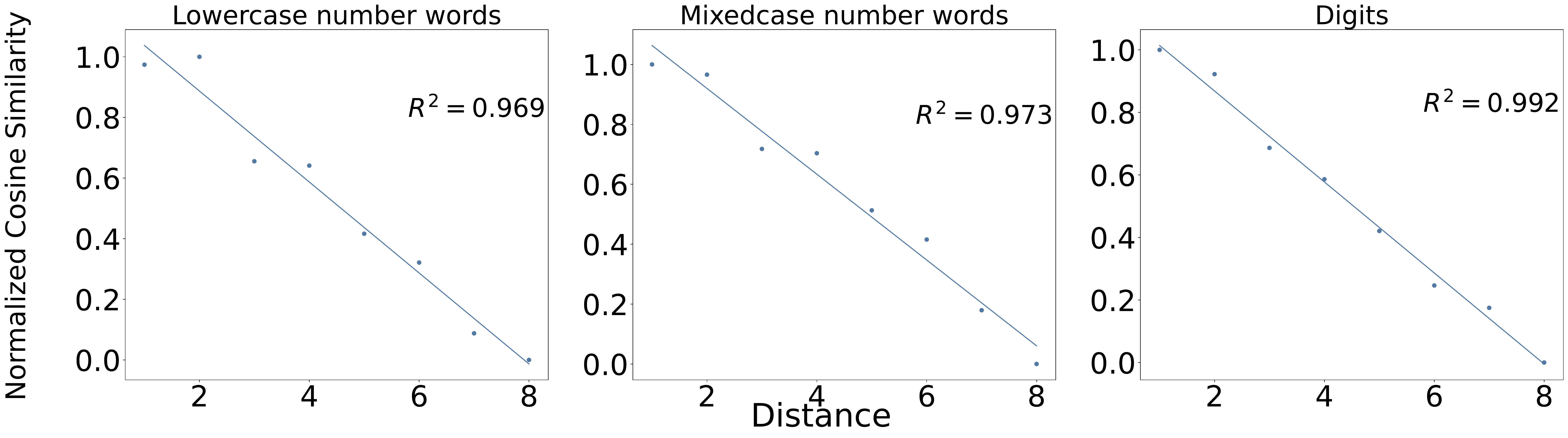}
\caption{Distance effect for the best-performing layer (9th layer) for the BART model}
\label{fig:distance_effect_bart_layer_8}
\end{figure*}

\renewcommand*{\MinNumber}{0.91}%
\renewcommand*{\MaxNumber}{0.97}%
\begin{table}[!h]
    \centering
    \resizebox{0.48\textwidth}{!}{%
    \begin{tabular}{cccccccR}
    \hline
        \multicolumn{1}{c}{\begin{tabular}[c]{@{}c@{}} Layer\end{tabular}} & T5 & BART & RoB & XLNET & \multicolumn{1}{c}{\begin{tabular}[c]{@{}c@{}} BERT\end{tabular}} & GPT-2 & \multicolumn{1}{c}{\begin{tabular}[C]{@{}c@{}} Avg. \end{tabular}} \\ \hline
        
        1 & 0.974 & 0.965 & 0.954 & 0.967 & 0.979 & 0.937 & 0.963 \\ 
        2 & 0.984 & 0.959 & 0.959 & 0.951 & 0.983 & 0.940 & 0.963 \\ 
        3 & 0.973 & 0.957 & 0.961 & 0.960 & 0.955 & 0.937 & 0.957 \\ 
        4 & 0.956 & 0.964 & 0.977 & 0.962 & 0.956 & 0.923 & 0.957 \\ 
        5 & 0.941 & 0.951 & 0.976 & 0.948 & 0.982 & 0.931 & 0.955 \\ 
        6 & 0.972 & 0.916 & 0.966 & 0.942 & 0.991 & 0.932 & 0.953 \\ 
        7 & 0.967 & 0.960 & 0.967 & 0.943 & 0.990 & 0.930 & 0.959 \\ 
        8 & 0.945 & 0.969 & 0.954 & 0.923 & 0.977 & 0.931 & 0.950 \\ 
        9 & 0.950 & 0.978 & 0.945 & 0.920 & 0.967 & 0.929 & 0.948 \\ 
        10 & 0.933 & 0.958 & 0.928 & 0.926 & 0.923 & 0.931 & 0.933 \\ 
        11 & 0.924 & 0.975 & 0.968 & 0.951 & 0.926 & 0.930 & 0.946 \\ 
        12 & 0.920 & 0.956 & 0.854 & 0.934 & 0.890 & 0.931 & 0.914 \\ \hline 
        
    \end{tabular}
    }
     \caption{Distance Effect: Averaged (across the three number formats) $R^2$ values of different LLMs for different layers when fitting a linear function. RoB: Roberta-base model, BERT: uncased variant.} 
\label{tab:distance_effect_model_layers}
\end{table}

\begin{table}[!ht]
    \centering
        \resizebox{0.41\textwidth}{!}{%

    \begin{tabular}{lcccc}
    \hline
    \multicolumn{1}{l}{\begin{tabular}[l]{@{}l@{}} LLMs$\backslash$Input\end{tabular}} 
         & \multicolumn{1}{c}{\begin{tabular}[c]{@{}c@{}}LC \end{tabular}}  & \multicolumn{1}{c}{\begin{tabular}[c]{@{}c@{}} MC\end{tabular}}  & \multicolumn{1}{c}{\begin{tabular}[c]{@{}c@{}} Digits \end{tabular}}  & \multicolumn{1}{c}{\begin{tabular}[c]{@{}c@{}} Avg. \end{tabular}} \\ \hline
        T5 & 0.986 & 0.937 & 0.936 & 0.953 \\ 
        BART & 0.942 & 0.951 & 0.983 & 0.959 \\ 
        RoBERTa & 0.945 & 0.943 & 0.964 & 0.951 \\ 
        XLNET & 0.888 & 0.965 & 0.979 & 0.944 \\ 
        BERT (uncased) & \multicolumn{2}{c}{0.976} & 0.944 & 0.960 \\ 
        GPT-2 & 0.906 & 0.904 & 0.986 & 0.932 \\ 
        \hline
        \multicolumn{1}{l}{\begin{tabular}[l]{@{}l@{}} Total Averages \\across models\end{tabular}}
         & 0.941 & 0.946 & \textbf{0.965} & 0.950 \\ \hline
    \end{tabular}
    }
    \caption{Distance Effect: Averaged (across layers) $R^2$ values of different LLMs on the three numbers when fitting a linear function. LC: Lowercase number words, MC: Mixed-case number words.}
\label{tab:distance_effect_models}
\end{table}

Recall that the distance effect is that people are slower (i.e., find it more difficult) to compare numbers the closer they are to each other on the MNL. We use the pipeline depicted in Figure \ref{fig:introduction0} to investigate if LLM representations are more similar to each other if the numbers are closer on the MNL. 

Evaluation of the distance effect in LLMs is done by fitting a straight line ($a+bx$) on the cosine similarity vs. distance plot. We first perform two operations on these cosine similarities: (1) We average the similarities across each distance (e.g., the point at distance 1 on the $x$-axis represents the average similarity of 1 vs. 2, 2 vs. 3, ..., 8 vs. 9). (2) We normalize the similarities to be in the range [0, 1]. These decisions allow relative output comparisons across different model architectures, which is not possible using the raw cosine similarities of each LLM. To illustrate model performance, the distance effects for the best-performing layer in terms of $R^2$ values for BART are shown in Figure \ref{fig:distance_effect_bart_layer_8} for the three number formats. The high $R^2$ values indicate a human-like distance effect. 
\renewcommand*{\MinNumber}{0.59}%
\renewcommand*{\MaxNumber}{0.72}%
\begin{table}[!h]
    \centering
    \resizebox{0.48\textwidth}{!}{%
    \begin{tabular}{cccccccR}
    \hline
        \multicolumn{1}{c}{\begin{tabular}[c]{@{}c@{}} Layer\end{tabular}} & T5 & BART & RoB & XLNET & \multicolumn{1}{c}{\begin{tabular}[c]{@{}c@{}} BERT\end{tabular}} & GPT-2 & \multicolumn{1}{c}{\begin{tabular}[c]{@{}c@{}} Avg.\end{tabular}} \\ \hline
        1 & 0.756 & 0.651 & 0.494 & 0.602 & 0.617 & 0.466 & 0.597 \\ 
        2 & 0.685 & 0.637 & 0.507 & 0.551 & 0.783 & 0.653 & 0.636 \\
        3 & 0.744 & 0.697 & 0.503 & 0.492 & 0.834 & 0.574 & 0.641 \\ 
        4 & 0.726 & 0.677 & 0.519 & 0.493 & 0.871 & 0.478 & 0.627 \\ 
        5 & 0.665 & 0.685 & 0.610 & 0.54 & 0.783 & 0.528 & 0.635 \\ 
        6 & 0.670 & 0.692 & 0.586 & 0.563 & 0.757 & 0.539 & 0.635 \\ 
        7 & 0.701 & 0.634 & 0.613 & 0.585 & 0.823 & 0.539 & 0.649 \\ 
        8 & 0.705 & 0.687 & 0.567 & 0.591 & 0.870 & 0.532 & 0.659 \\ 
        9 & 0.697 & 0.757 & 0.581 & 0.566 & 0.877 & 0.541 & 0.670 \\ 
        10 & 0.727 & 0.694 & 0.622 & 0.555 & 0.905 & 0.533 & 0.672 \\ 
        11 & 0.729 & 0.756 & 0.734 & 0.602 & 0.911 & 0.547 & 0.713 \\ 
        12 & 0.703 & 0.702 & 0.744 & 0.662 & 0.889 & 0.550 & 0.708 \\ \hline
    \end{tabular}
    }
     \caption{Size Effect: Averaged (across inputs) $R^2$ values of different LLMs on different input layers when fitting a linear function. RoB: Roberta-base model, BERT: uncased variant.}
\label{tab:size_effect_model_layers}
\end{table}

All of the models show strong distance effects for all layers, as shown in Table \ref{tab:distance_effect_model_layers}, and for all number formats, as shown in Table \ref{tab:distance_effect_models}. Interestingly, LLMs are less likely to reveal the distance effect as layer count increases (Table \ref{tab:distance_effect_model_layers}). For example, layer one results in the strongest distance effect while layer twelve is the least representative of the distance effect. With respect to number format, passing \emph{digits} as inputs tended to produce stronger distance effects than passing number words (Table \ref{tab:distance_effect_models}); this pattern was present for four of the six LLMs (i.e., all but T5 and BERT). 

\subsection{The Size Effect}

\begin{figure*}[!ht]
\centering
\includegraphics[trim = {5.5cm 1cm 0cm 1cm}, width=0.81\textwidth]{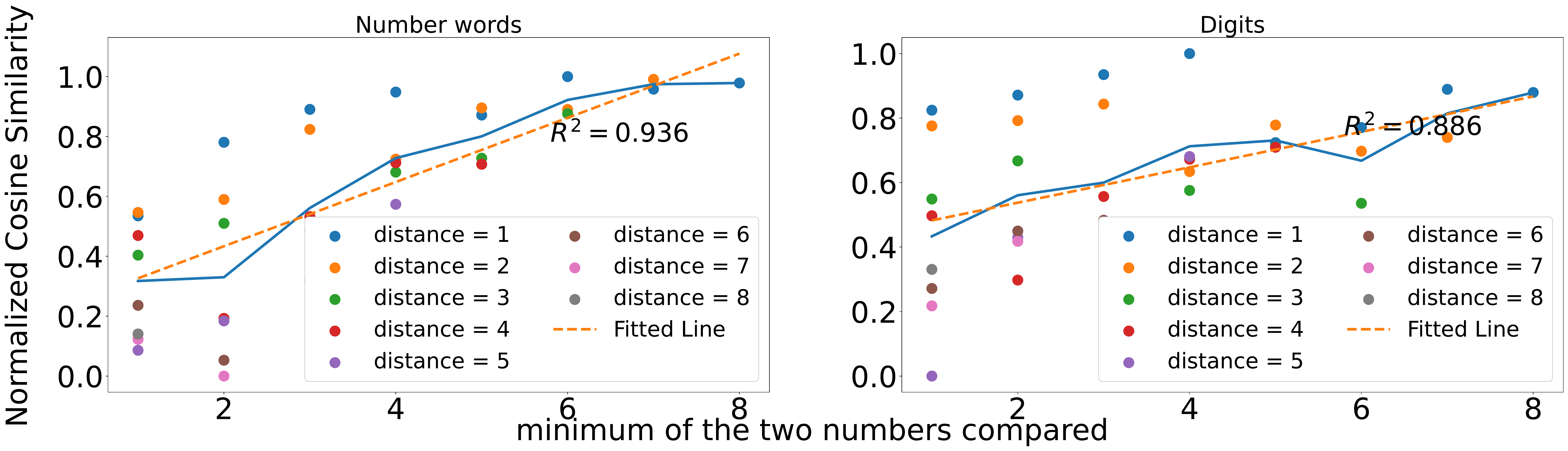}
\caption{Size effect for the best-performing layer for the BERT model (layer 11).}
\label{fig:size_bert_layer_10}
\end{figure*}

The size effect holds for comparisons of the same distance (e.g., for a distance of 1, these include 1 vs. 2, 2 vs. 3, ..., 8 vs. 9). Among these comparisons, those involving larger numbers (e.g., 8 vs. 9) are made more slowly (i.e., people find them more difficult) than those involving smaller numbers (e.g., 1 vs. 2). That larger numbers are harder to differentiate than smaller numbers aligns with the logarithmically compressed MNL depicted in Figure \ref{fig:introduction0}d. This study evaluates whether a given LLM shows a size effect on a given layer for numbers of a given format by plotting the normalized cosine similarities against the size of the comparison, defined as the minimum of the two numbers being compared. For each minimum value (points on the $x$-axis), we average the similarities for all comparisons to form a single point (vertical compression). We then fit a straight line $(ax+b)$ over the vertically compressed averages (blue line in Figure \ref{fig:size_bert_layer_10}) to obtain the $R^2$ values (scores). To illustrate model performance, the size effects for the best-performing layer of the BERT-uncased model (in terms of $R^2$ values) are shown in Figure \ref{fig:size_bert_layer_10}.
Similar to the results for the distance effect, the high $R^2$ values indicate a human-like size effect.

Interestingly, Table \ref{tab:size_effect_model_layers} generally shows an increasing trend in the layer-wise capability of capturing the size effect across the six LLMs. This is opposite to the trend observed across layers for the distance effect. Table \ref{tab:size_effect_models} shows that using digits as the input values yields significantly better $R^2$ values than the other number formats. In fact, this is the only number format for which the models produce strong size effects. However, the vertical compression of points fails to capture the spread of points across the $y$-axis for each point on the $x$-axis. This spread, a limitation of the size effect analysis, is captured in the ratio effect (section \ref{ratio_effect}).

\begin{table}[!ht]
    \centering
    \resizebox{0.41\textwidth}{!}{%
    \begin{tabular}{lcccc}
    \hline
    \multicolumn{1}{l}{\begin{tabular}[l]{@{}l@{}} LLMs$\backslash$Input\end{tabular}} 
         & \multicolumn{1}{c}{\begin{tabular}[c]{@{}c@{}} LC \end{tabular}}  & \multicolumn{1}{c}{\begin{tabular}[c]{@{}c@{}} MC \end{tabular}}  & \multicolumn{1}{c}{\begin{tabular}[c]{@{}c@{}} Digits \end{tabular}}  & \multicolumn{1}{c}{\begin{tabular}[c]{@{}c@{}} Avg. \end{tabular}} \\ \hline
       T5 & 0.702 & 0.539 & 0.886 & 0.709 \\ 
        BART & 0.614 & 0.568 & 0.885 & 0.689 \\ 
        RoBERTa & 0.520 & 0.466 & 0.783 & 0.59 \\ 
        XLNET & 0.500 & 0.408 & 0.793 & 0.567 \\ 
        BERT (uncased) & \multicolumn{2}{c}{0.803} & 0.851 & 0.827 \\ 
        GPT-2 & 0.434 & 0.332 & 0.853 & 0.54 \\ \hline
        \multicolumn{1}{l}{\begin{tabular}[l]{@{}l@{}} Total Averages \\across models\end{tabular}} & 0.596 & 0.519 & \textbf{0.842} & 0.654 \\ \hline
    \end{tabular}
    }
    \caption{Size Effect: Averaged (across layers) $R^2$ values of different LLMs on the three number formats when fitting a linear function. LC: Lowercase number words, MC: Mixed-case number words.}
\label{tab:size_effect_models}
\end{table}

\subsection{The Ratio Effect}
\label{ratio_effect}
\begin{table}[!ht]
    \centering
    \resizebox{0.41\textwidth}{!}{%
    \begin{tabular}{lcccc}
    \hline
    \multicolumn{1}{l}{\begin{tabular}[l]{@{}l@{}} LLMs$\backslash$Input\end{tabular}} 
         & \multicolumn{1}{c}{\begin{tabular}[c]{@{}c@{}} LC \end{tabular}}  & \multicolumn{1}{c}{\begin{tabular}[c]{@{}c@{}} MC \end{tabular}}  & \multicolumn{1}{c}{\begin{tabular}[c]{@{}c@{}} Digits\end{tabular}}  & \multicolumn{1}{c}{\begin{tabular}[c]{@{}c@{}} Avg. \end{tabular}} \\ \hline
        T5 & 0.852 & 0.756 & 0.868 & 0.826 \\ 
        BART & 0.786 & 0.833 & 0.897 & 0.838 \\ 
        RoBERTa & 0.714 & 0.747 & 0.746 & 0.736 \\ 
        XLNET & 0.729 & 0.761 & 0.901 & 0.797 \\ 
        BERT (uncased) & \multicolumn{2}{c}{0.906} & 0.757 & 0.831 \\ 
        GPT-2 & 0.686 & 0.758 & 0.681 & 0.709 \\ \hline
        \multicolumn{1}{l}{\begin{tabular}[l]{@{}l@{}} Total Averages\\across models\end{tabular}}
         & 0.779 & 0.793 & \textbf{0.808} & 0.789 \\ \hline
    \end{tabular}
    }

    \caption{Ratio Effect: Averaged (across layers) $R^2$ values of different LLMs on different number formats when fitting a negative exponential function. LC: Lowercase number words, MC: Mixed-case number words.}
\label{tab:ratio_effect_models}
\end{table}

\renewcommand*{\MinNumber}{0.69}%
\renewcommand*{\MaxNumber}{0.82}%
\begin{table}[!ht]
    \centering
    \resizebox{0.48\textwidth}{!}{%
    \begin{tabular}{cccccccR}
    \hline
        \multicolumn{1}{c}{\begin{tabular}[c]{@{}c@{}} Layer\end{tabular}} & T5 & BART & RoB & XLNET & \multicolumn{1}{c}{\begin{tabular}[c]{@{}c@{}} BERT\end{tabular}} & GPT-2 & \multicolumn{1}{c}{\begin{tabular}[c]{@{}c@{}}Avg.\end{tabular}} \\ \hline
        1 & 0.850 & 0.820 & 0.756 & 0.868 & 0.837 & 0.735 & 0.811 \\ 
        2 & 0.865 & 0.837 & 0.745 & 0.828 & 0.878 & 0.755 & 0.819 \\ 
        3 & 0.846 & 0.861 & 0.725 & 0.820 & 0.853 & 0.738 & 0.807 \\ 
        4 & 0.847 & 0.859 & 0.739 & 0.822 & 0.820 & 0.659 & 0.791 \\
        5 & 0.851 & 0.847 & 0.805 & 0.825 & 0.847 & 0.695 & 0.812 \\ 
        6 & 0.880 & 0.821 & 0.800 & 0.816 & 0.883 & 0.703 & 0.817 \\ 
        7 & 0.867 & 0.811 & 0.795 & 0.810 & 0.883 & 0.698 & 0.811 \\ 
        8 & 0.824 & 0.849 & 0.780 & 0.780 & 0.880 & 0.702 & 0.803 \\ 
        9 & 0.806 & 0.852 & 0.780 & 0.746 & 0.861 & 0.705 & 0.791 \\ 
        10 & 0.785 & 0.821 & 0.720 & 0.754 & 0.779 & 0.704 & 0.760 \\ 
        11 & 0.755 & 0.849 & 0.666 & 0.781 & 0.769 & 0.702 & 0.754 \\ 
        12 & 0.731 & 0.834 & 0.516 & 0.717 & 0.687 & 0.708 & 0.699 \\ \hline
    \end{tabular}
    }
     \caption{Ratio Effect: Averaged (across number formats) $R^2$ values of different LLMs on different input layers when fitting a negative exponential function. RoB: Roberta-base model, BERT: uncased variant.}
\label{tab:ratio_effect_model_layers}
\end{table}

\begin{figure*}[!h]
\centering
\includegraphics[trim = {5.5cm 1cm 0cm 1cm}, width=0.82\textwidth]{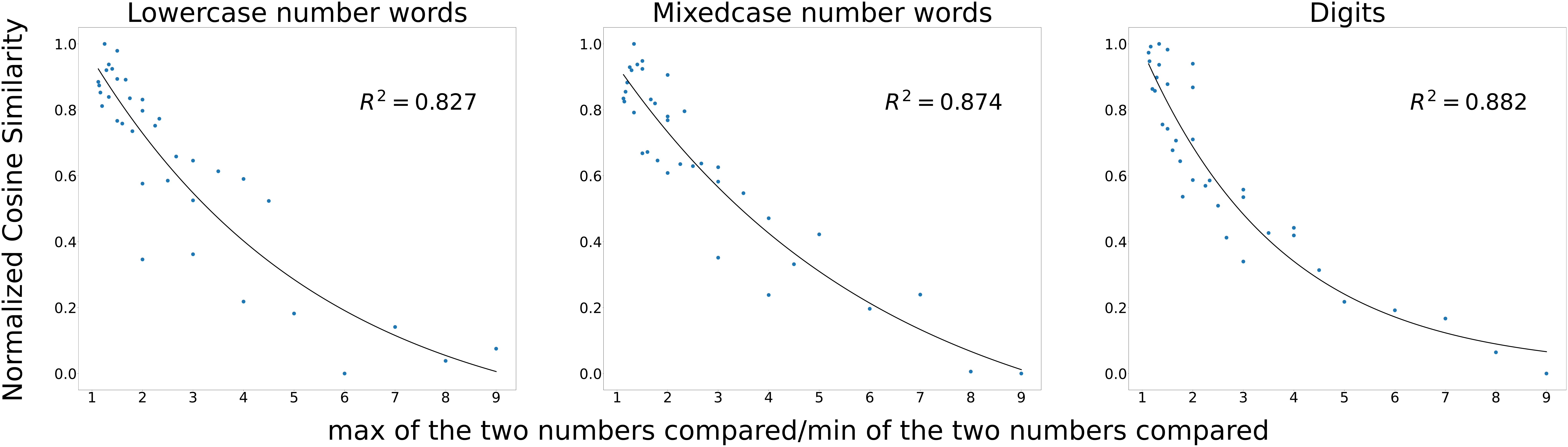}
\caption{Ratio effect for the best-performing layer for the BART model (layer 3).}
\label{fig:ratio_bart_layer_2}
\end{figure*}
The ratio effect in humans can be thought of as simultaneously capturing both the distance and size effects. Behaviorally, the time to compare $x$ vs. $y$ is a decreasing function of the ratio of the larger number over the smaller number, i.e., of $\frac{\max(x, y)}{\min(x, y)}$. In fact, the function is nonlinear as depicted in Figure \ref{fig:introduction0}c. For the LLMs, we plot the normalized cosine similarity vs. $\frac{\max(x, y)}{\min(x, y)}$. To each plot, we fit the negative exponential function $a*e^{-bx}+c$ and evaluate the resulting $R^2$. To illustrate model performance, Figure \ref{fig:ratio_bart_layer_2} shows the ratio effects for the best-fitting layer of the BART model for the three number formats.
As observed with the distance and size effect, the high $R^2$ values of the LLMs indicate a human-like ratio effect in the models. 

\begin{table}[!ht]
    \centering
    \resizebox{0.41\textwidth}{!}{%
    \begin{tabular}{lcccc}
    \hline
    \multicolumn{1}{l}{\begin{tabular}[l]{@{}l@{}} LLMs$\backslash$Input\end{tabular}} 
         & \multicolumn{1}{c}{\begin{tabular}[c]{@{}c@{}} LC \end{tabular}}  & \multicolumn{1}{c}{\begin{tabular}[c]{@{}c@{}} MC\end{tabular}}  & \multicolumn{1}{c}{\begin{tabular}[c]{@{}c@{}} Digits\end{tabular}}  & \multicolumn{1}{c}{\begin{tabular}[c]{@{}c@{}} Avg. \end{tabular}} \\ \hline
       T5 & 0.489 & 0.526 & 0.410 & 0.475 \\ 
        BART & 0.676 & 0.714 & 0.678 & 0.690 \\ 
        RoBERTa & 0.520 & 0.597 & 0.587 & 0.568 \\ 
        XLNET & 0.622 & 0.620 & 0.622 & 0.621 \\ 
        BERT (uncased) & \multicolumn{2}{c}{0.312} & 0.423 & 0.368 \\ 
        GPT-2 & 0.566 & 0.513 & \textbf{0.828} & 0.636 \\ \hline
        \multicolumn{1}{l}{\begin{tabular}[l]{@{}l@{}} Total Averages \\across models\end{tabular}} & 0.531 & 0.547 & \textbf{0.591} & 0.560 \\ \hline
    \end{tabular}
    }
    \caption{Averaged (across layers) correlations when comparing MDS values with $Log_{10}1$ to $Log_{10}9$ for different LLMs. LC: Lowercase number words, MC: Mixed-case number words.}
\label{tab:mds}
\end{table}

\renewcommand*{\MinNumber}{0.38}%
\renewcommand*{\MaxNumber}{0.64}%

\begin{table}[!h]
    \centering
    \resizebox{0.48\textwidth}{!}{%
    \begin{tabular}{cccccccR}
    \hline
        \multicolumn{1}{l}{\begin{tabular}[l]{@{}l@{}} Layer\end{tabular}} & T5 & BART & RoB & XLNET & \multicolumn{1}{c}{\begin{tabular}[c]{@{}c@{}} BERT\end{tabular}} & GPT-2 & \multicolumn{1}{c}{\begin{tabular}[c]{@{}c@{}} Avg.\end{tabular}} \\ \hline
       1 & 0.686 & 0.679 & 0.602 & 0.595 & 0.739 & 0.526 & 0.638 \\ 
        2 & 0.271 & 0.693 & 0.763 & 0.734 & 0.704 & 0.669 & 0.639 \\ 
        3 & 0.374 & 0.657 & 0.772 & 0.704 & 0.456 & 0.685 & 0.608 \\ 
        4 & 0.385 & 0.728 & 0.489 & 0.621 & 0.425 & 0.663 & 0.552 \\ 
        5 & 0.476 & 0.733 & 0.597 & 0.707 & 0.448 & 0.615 & 0.596 \\ 
        6 & 0.540 & 0.739 & 0.571 & 0.598 & 0.465 & 0.608 & 0.587 \\ 
        7 & 0.687 & 0.696 & 0.250 & 0.677 & 0.445 & 0.665 & 0.570 \\ 
        8 & 0.529 & 0.624 & 0.594 & 0.591 & 0.189 & 0.624 & 0.525 \\ 
        9 & 0.544 & 0.718 & 0.691 & 0.566 & 0.400 & 0.671 & 0.598 \\ 
        10 & 0.502 & 0.624 & 0.697 & 0.563 & 0.394 & 0.613 & 0.566 \\ 
        11 & 0.195 & 0.708 & 0.602 & 0.543 & -0.013 & 0.675 & 0.451 \\ 
        12 & 0.509 & 0.677 & 0.186 & 0.557 & -0.239 & 0.615 & 0.384 \\ \hline
    \end{tabular}
    }
     \caption{Averaged (across inputs) correlations of different LLMs on different model layers when comparing MDS values with $Log_{10}1$ to $Log_{10}9$. RoB: Roberta-base model, BERT: uncased variant.}
\label{tab:mds_model_layers}
\end{table}

\renewcommand*{\MinNumber}{0.01}%
\renewcommand*{\MaxNumber}{0.25}%
\begin{table}[!ht]
    \centering
    \resizebox{0.46\textwidth}{!}{%
    \begin{tabular}{cccccccc}
    \hline
        \multicolumn{1}{l}{\begin{tabular}[l]{@{}l@{}} Number\end{tabular}} & T5 & BART & RoB & XLNET & \multicolumn{1}{c}{\begin{tabular}[c]{@{}c@{}} BERT \end{tabular}} & GPT-2 & \multicolumn{1}{c}{\begin{tabular}[c]{@{}c@{}} Avg.\end{tabular}} \\ \hline
        1 & 0.01 & 0.00 & 0.02 & 0.00 & 0.02 & 0.00 & 0.01 \\ 
        2 & 0.10 & 0.17 & 0.15 & 0.17 & 0.09 & 0.12 & \textbf{0.13} \\ 
        3 & 0.07 & 0.05 & 0.07 & 0.10 & 0.06 & 0.10 & 0.07 \\ 
        4 & 0.05 & 0.04 & 0.05 & 0.05 & 0.03 & 0.05 & 0.04 \\ 
        5 & 0.17 & 0.09 & 0.07 & 0.05 & 0.20 & 0.05 & \textbf{0.11} \\ 
        6 & 0.02 & 0.04 & 0.08 & 0.02 & 0.06 & 0.04 & 0.04 \\ 
        7 & 0.09 & 0.08 & 0.11 & 0.04 & 0.20 & 0.06 & \textbf{0.10} \\ 
        8 & 0.04 & 0.01 & 0.08 & 0.01 & 0.09 & 0.05 & 0.05 \\ 
        9 & 0.40 & 0.08 & 0.17 & 0.18 & 0.44 & 0.17 & \textbf{0.24} \\ 
        \hline
    \end{tabular}
    }
    
     \caption{Residual analysis on MDS outputs in 1 dimension on the base variants of the model.  RoB: Roberta-base model, BERT: uncased variant.}
\label{tab:mds_residual}
\end{table}

\subsection{Multidimensional Scaling}

Along with the three magnitude effects, we also investigate whether the number representations of LLMs are consistent with the human MNL. To do so, we utilize multidimensional scaling \cite{mds_BorgGroenen2005, mds_Ding2018}. MDS offers a method for recovering the latent structure in the matrix of cosine (dis)similarities between the vector representations of all pairs of numbers (for a given LLM, layer, and number format). It arranges each number in a space of $N$ dimensions such that the distance between each pair of points is consistent with the cosine dissimilarity between their vector representations.

We fix $N = 1$ to recover the latent MNL representation for each LLM, layer, and number format. For each solution, we anchor the point for "1" to the left side and evaluate whether the resulting visualization approximates the log compressed MNL as shown in Figure \ref{fig:introduction0}d. To quantify this approximation, we calculate the correlation between the positions of the numbers 1 to 9 in the MDS solution and the expected values (log(1) to log (9)) of the human MNL; see Table \ref{tab:mds}. All inputs have similar correlation values. Surprisingly, GPT-2 with digits as the number format (and averaged across all layers) shows a considerably higher correlation with the log-compressed MNL than all other models and number formats. The average correlation between latent model number lines and the log compressed MNL decreases over the 12 layers; see Table \ref{tab:mds_model_layers}.

\begin{figure}[!ht]
\centering
\includegraphics[trim = {5cm 4cm 0cm 1cm}, width=0.41\textwidth]{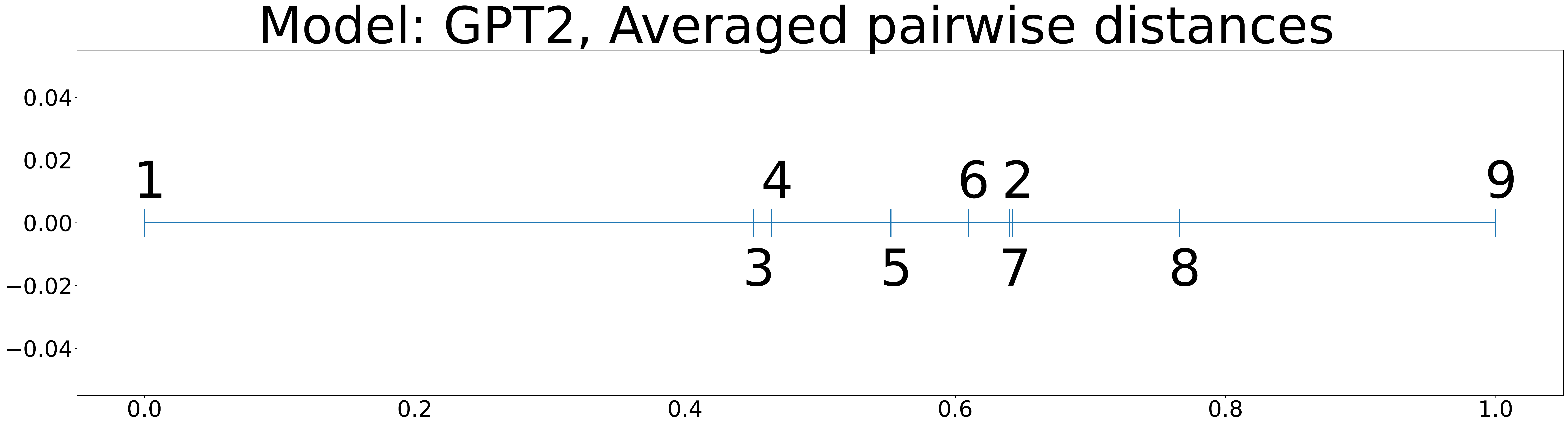}
\caption{MDS visualization on averaged distances of the GPT-2 model for all number formats and layers.}
\label{fig:mds_gpt_2_agg}
\end{figure}

We visualize the latent number line of GPT-2 by averaging the cosine dissimilarity matrix across layers and number formats, submitting this to MDS, and requesting a one-dimensional solution; see Figure \ref{fig:mds_gpt_2_agg}. This representation shows some evidence of log compression, though with a few exceptions. One obvious exception is the right displacement of 2 away from 1. Another is the right displacement of 9 very far from 8. 

To better understand if this is a statistical artifact of GPT-2 or a more general difference between number understanding in humans versus LLMs, we perform a residual analysis comparing positions on the model's number line to those on the human MNL. We choose the digits number format, estimate the latent number line representation averaged across the layers of each model, and compute the residual between the position of each number in this representation compared to the human MNL. This analysis is presented in Table \ref{tab:mds_residual}. For 1, all models show a residual value of less than 0.03. This makes sense given our decision to anchor the latent number lines to 1 on the left side. The largest residuals are for 2 and 9, consistent with the anomalies noticed for the GPT-2 solution in Figure \ref{fig:mds_gpt_2_agg}. These anomalies are a target for future research. We note here that 2 is often privileged even in languages such as Piraha and Mundurucu that have very limited number of word inventories\cite{cite1, cite2}. Further note that 9 has special significance as a ``bargain price numeral'' in many cultures, a fact that is often linguistically marked \cite{cite3}.

\subsection{Ablation studies: Base vs Large Model Variants}
We investigate changes in model behaviors when increasing the number of parameters for the same architectures. We use the larger variants of each of the LLMs listed in Table \ref{tab:models}. The detailed tabular results of the behaviors are presented in Appendix section \ref{variants}; see Tables \ref{tab:inputs_large}, \ref{tab:mds_residual_large}, and \ref{tab:layers_large}. Here, we summarize key takeaways from the ablation studies: 
\begin{compactitem}
    \item The distance and ratio effects of the large variants of models \emph{align with human performance characteristics}. Similar to the results for the base variants, the size effect is only observed when the input type is digits.
    \item We observe the \emph{same decreasing trend} in the layer-wise capability of capturing the \emph{distance effect, ratio effect, and the MDS correlation values} in the Large variants of LLMs as observed in the base variants. The increasing trend in the layer-wise capability of the size effect is \emph{not} observed in the Larger LLMs.
    \item Residual analysis shows high deviation for the numbers "2", "5", and "9"; which is \emph{in line} with our observations for the base variations.
\end{compactitem}

\section{Conclusion}
This paper investigates the performance characteristics in various LLMs across numerous configurations, looking for three number-magnitude comparison effects: distance, size, and ratio.  Our results show that LLMs show human-like distance and ratio effects across number formats.
The size effect is also observed among models for the digit number format, but not for the other number formats, showing that LLMs do not completely capture numeration. Using MDS to scale down the pairwise (dis)similarities between number representations produces varying correspondences between LLMs and the logarithmically compressed MNL of humans, with GPT-2 showing the highest correlation (using digits as inputs). Our residual analysis exhibits high deviation from expected outputs for the numbers 2, 5, 9 which we explain through patterns observed in previous linguistics studies. The behavioral benchmarking of the numeric magnitude representations of LLMs presented here helps us understand the cognitive plausibility of the representations the models learn. Our results show that LLM pre-training allows models to approximately learn human-like behaviors for two out of the three magnitude effects without the need to posit explicit neural circuitry. Future work on building pre-trained architectures to improve numerical cognition abilities should also be evaluated using these three effects. 
\section{Limitations}
Limitations to our work are as follows: (1) We only study the three magnitude effects for the number word and digit denotations of the numbers 1 to 9. The effects for the number 0, numbers greater than 10, decimal numbers, negative numbers, etc. are beyond the scope of this study. Future work can design behavioral benchmark for evaluating whether LLMs shows these effects for these other number classes. (2) The mapping of LLM behaviors to human behaviors and effects might vary for each effect. Thus, we might require a different linking hypothesis for each such effect. (3) We only use the models built for English tasks and do not evaluate multi-lingual models. (4) We report and analyze aggregated scores across different dimensions. There can be some information loss in this aggregation. (5) Our choice of models is limited by certain resource constraints. Future works can explore the use of other foundation / super-large models (1B parameters +) and API-based models like GPT3 and OPT3. (6) The behavioral analysis of this study is one-way: we look for human performance characteristics and behaviors in LLMs. Future research can utilize LLMs to discover new numerical effects and look for the corresponding performance characteristics in humans. This could spur new research in cognitive science. (7) The results show similar outputs to low dimensional human output and show that we do not need explicit neural circuitry for number understanding. We do not suggest models actually are humanlike in \textit{how} they process numbers.
\bibliography{anthology,custom}
\bibliographystyle{acl_natbib}

\appendix

\section{Appendix}
\label{sec:appendix}

\subsection{Variants in Large Language Models}
\label{variants}

\begin{table}[!ht]
    \centering
    \resizebox{0.48\textwidth}{!}{%
    \begin{tabular}{lcccc}
    \hline
       \multicolumn{1}{l}{\begin{tabular}[c]{@{}c@{}} Inputs$\backslash$Effects\end{tabular}}  & 
       \multicolumn{1}{l}{\begin{tabular}[c]{@{}c@{}} Averaged \\ Distance \\ Effect\end{tabular}} 
       
        & \multicolumn{1}{l}{\begin{tabular}[c]{@{}c@{}} Averaged \\ Size \\ Effect\end{tabular}} 
       
        &\multicolumn{1}{l}{\begin{tabular}[c]{@{}c@{}} Averaged\\ Ratio\\ Effect\end{tabular}} 
       
        & 
       \multicolumn{1}{l}{\begin{tabular}[c]{@{}c@{}} Averaged MDS\\ Correlation\\ values\end{tabular}} 
       
        \\ \hline
         \multicolumn{1}{l}{\begin{tabular}[l]{@{}l@{}} Lowercase\\ number words\end{tabular}} & 0.909 & 0.587 & 0.730 & 0.593 \\ 
        \multicolumn{1}{l}{\begin{tabular}[l]{@{}l@{}} Mixedcase\\ number words\end{tabular}} & 0.933 & 0.514 & 0.749 & 0.460 \\ 
        Digits & 0.930 & 0.678 & 0.707 & 0.548 \\ 
        \multicolumn{1}{c}{\begin{tabular}[c]{@{}c@{}} Total \\Averages\end{tabular}} & 0.927 & 0.595 & 0.727 & 0.534 \\ \hline
    \end{tabular}
    }
    \caption{Averaged distance effect, size effect, ratio effect, and the MDS correlation values for the different input types of the models.}
    \label{tab:inputs_large}
\end{table}

\begin{table}[!ht]
    \centering
    \resizebox{0.46\textwidth}{!}{%
    \begin{tabular}{cccccccc}
    \hline
        \multicolumn{1}{l}{\begin{tabular}[l]{@{}l@{}} Number\end{tabular}} & T5 & BART & RoB & XLNET & \multicolumn{1}{c}{\begin{tabular}[c]{@{}c@{}} BERT\end{tabular}} & GPT-2 & \multicolumn{1}{c}{\begin{tabular}[c]{@{}c@{}} Avg.\end{tabular}} \\ \hline
        1 & 0.04 & 0.01 & 0.01 & 0.01 & 0.01 & 0.00 & 0.01 \\ 
        2 & 0.09 & 0.17 & 0.09 & 0.16 & 0.07 & 0.12 & \textbf{0.12} \\ 
        3 & 0.02 & 0.09 & 0.04 & 0.07 & 0.03 & 0.10 & 0.06 \\ 
        4 & 0.02 & 0.07 & 0.03 & 0.04 & 0.03 & 0.07 & 0.04 \\ 
        5 & 0.12 & 0.07 & 0.13 & 0.17 & 0.16 & 0.02 & \textbf{0.11} \\ 
        6 & 0.20 & 0.06 & 0.06 & 0.05 & 0.10 & 0.02 & 0.08 \\ 
        7 & 0.17 & 0.09 & 0.09 & 0.07 & 0.12 & 0.02 & 0.09 \\ 
        8 & 0.22 & 0.09 & 0.05 & 0.06 & 0.09 & 0.03 & 0.09 \\ 
        9 & 0.15 & 0.19 & 0.25 & 0.36 & 0.25 & 0.14 & \textbf{0.22} \\ 
        \hline
    \end{tabular}
    }
    
     \caption{Residual analysis on MDS outputs in 1 dimension on the large variants of the models. RoB: Roberta-base model, BERT: uncased variant.}
\label{tab:mds_residual_large}
\end{table}

\renewcommand*{\MinNumber}{0.410}%
\renewcommand*{\MaxNumber}{0.98}%
\begin{table}[!ht]
    \centering
    \resizebox{0.48\textwidth}{!}{%
    \begin{tabular}{ccccc}
    \hline
       \multicolumn{1}{l}{\begin{tabular}[c]{@{}c@{}} Layer$\backslash$Effects\end{tabular}}  & 
       \multicolumn{1}{l}{\begin{tabular}[c]{@{}c@{}} Averaged \\ Distance \\ Effect\end{tabular}} 
       
        & \multicolumn{1}{l}{\begin{tabular}[c]{@{}c@{}} Averaged \\ Size \\ Effect\end{tabular}} 
       
        &\multicolumn{1}{l}{\begin{tabular}[c]{@{}c@{}} Averaged\\ Ratio\\ Effect\end{tabular}} 
       
        & 
       \multicolumn{1}{l}{\begin{tabular}[c]{@{}c@{}} Averaged MDS\\ Correlation\\ values\end{tabular}} 
       
        \\ \hline
        1 & 0.967 & 0.647 & 0.825 & 0.643 \\ 
        2 & 0.963 & 0.549 & 0.718 & 0.557 \\ 
        3 & 0.964 & 0.587 & 0.736 & 0.584 \\ 
        4 & 0.968 & 0.622 & 0.765 & 0.544 \\ 
        5 & 0.962 & 0.632 & 0.763 & 0.423 \\ 
        6 & 0.958 & 0.641 & 0.774 & 0.483 \\ 
        7 & 0.957 & 0.591 & 0.752 & 0.526 \\ 
        8 & 0.956 & 0.608 & 0.753 & 0.550 \\ 
        9 & 0.956 & 0.599 & 0.773 & 0.625 \\ 
        10 & 0.944 & 0.612 & 0.766 & 0.610 \\ 
        11 & 0.938 & 0.608 & 0.742 & 0.526 \\ 
        12 & 0.923 & 0.604 & 0.726 & 0.557 \\ 
        13 & 0.939 & 0.659 & 0.739 & 0.538 \\ 
        14 & 0.944 & 0.656 & 0.755 & 0.562 \\ 
        15 & 0.940 & 0.645 & 0.751 & 0.500 \\ 
        16 & 0.933 & 0.611 & 0.741 & 0.509 \\ 
        17 & 0.934 & 0.567 & 0.730 & 0.550 \\ 
        18 & 0.933 & 0.580 & 0.723 & 0.505 \\ 
        19 & 0.919 & 0.559 & 0.690 & 0.527 \\ 
        20 & 0.900 & 0.557 & 0.671 & 0.535 \\ 
        21 & 0.867 & 0.558 & 0.644 & 0.553 \\ 
        22 & 0.854 & 0.571 & 0.664 & 0.524 \\ 
        23 & 0.829 & 0.509 & 0.633 & 0.484 \\ 
        24 & 0.805 & 0.508 & 0.622 & 0.414 \\ \hline
    \end{tabular}
    }
    \caption{Averaged distance effect, size effect, ratio effect, and MDS correlation values for the 24 layers of the models.}
    \label{tab:layers_large}
\end{table}
          

\begin{table}[!ht]
    \centering
        \resizebox{0.41\textwidth}{!}{%

    \begin{tabular}{lcccc}
    \hline
    \multicolumn{1}{l}{\begin{tabular}[l]{@{}l@{}} LLMs$\backslash$Input\end{tabular}} 
         & \multicolumn{1}{c}{\begin{tabular}[c]{@{}c@{}} LC \end{tabular}}  & \multicolumn{1}{c}{\begin{tabular}[c]{@{}c@{}} MC\end{tabular}}  & \multicolumn{1}{c}{\begin{tabular}[c]{@{}c@{}} Digits \end{tabular}}  & \multicolumn{1}{c}{\begin{tabular}[c]{@{}c@{}} Avg. \end{tabular}} \\ \hline
        T5 & 0.961 & 0.957 & 0.974 & 0.964 \\ 
        BART & 0.892 & 0.957 & 0.845 & 0.898 \\ 
        RoBERTa & 0.893 & 0.959 & 0.946 & 0.933 \\ 
        XLNET & 0.924 & 0.952 & 0.855 & 0.910 \\ 
        BERT (uncased) & \multicolumn{2}{c}{0.837} & 0.969 & 0.903 \\ 
        GPT-2 & 0.946 & 0.934 & 0.987 & 0.956 \\ 
        \hline
        \multicolumn{1}{l}{\begin{tabular}[l]{@{}l@{}} Total Averages \\across models\end{tabular}}
         & 0.909 & 0.933 & 0.930 & 0.927 \\ \hline
    \end{tabular}
    }
    \caption{Distance Effect: Averaged (across layers) $R^2$ values of different \emph{Larger variants of} LLMs on different input types when fitting a linear function. LC: Lowercase number words, MC: Mixedcase number words.}
\label{tab:distance_effect_models_large}
\end{table}

\begin{table}[!ht]
    \centering
        \resizebox{0.41\textwidth}{!}{%

    \begin{tabular}{lcccc}
    \hline
    \multicolumn{1}{l}{\begin{tabular}[l]{@{}l@{}} LLMs$\backslash$Input\end{tabular}} 
         & \multicolumn{1}{c}{\begin{tabular}[c]{@{}c@{}}LC \end{tabular}}  & \multicolumn{1}{c}{\begin{tabular}[c]{@{}c@{}} MC\end{tabular}}  & \multicolumn{1}{c}{\begin{tabular}[c]{@{}c@{}} Digits\end{tabular}}  & \multicolumn{1}{c}{\begin{tabular}[c]{@{}c@{}} Avg. \end{tabular}} \\ \hline
        T5 & 0.720 & 0.730 & 0.840 & 0.763 \\ 
        BART & 0.697 & 0.644 & 0.380 & 0.574 \\ 
        RoBERTa & 0.468 & 0.267 & 0.677 & 0.471 \\ 
        XLNET & 0.533 & 0.448 & 0.510 & 0.497 \\ 
        BERT (uncased) & \multicolumn{2}{c}{0.635} & 0.712 & 0.674 \\ 
        GPT-2 & 0.467 & 0.358 & 0.950 & 0.592 \\ 
        \hline
        \multicolumn{1}{l}{\begin{tabular}[l]{@{}l@{}} Total Averages \\across models\end{tabular}}
         & 0.587 & 0.514 & 0.678 & 0.595 \\ \hline
    \end{tabular}
    }
    \caption{Size Effect: Averaged (across layers) $R^2$ values of different \emph{Larger variants of} LLMs on different input types when fitting a linear function. LC: Lowercase number words, MC: Mixedcase number words.}
\label{tab:size_effect_models_large}
\end{table}

\renewcommand*{\MinNumber}{0.80}%
\renewcommand*{\MaxNumber}{0.97}%

\begin{table}[!h]
    \centering
    \resizebox{0.48\textwidth}{!}{%
    \begin{tabular}{cccccccR}
    \hline
        \multicolumn{1}{c}{\begin{tabular}[c]{@{}c@{}} Layer\end{tabular}} & T5 & BART & RoB & XLNET & \multicolumn{1}{c}{\begin{tabular}[c]{@{}c@{}} BERT\end{tabular}} & GPT-2 & \multicolumn{1}{c}{\begin{tabular}[C]{@{}c@{}} Avg.\end{tabular}} \\ \hline
           1 & 0.978 & 0.948 & 0.968 & 0.972 & 0.978 & 0.959 & 0.967 \\ 
        2 & 0.977 & 0.958 & 0.962 & 0.976 & 0.967 & 0.940 & 0.963 \\ 
        3 & 0.977 & 0.970 & 0.931 & 0.979 & 0.977 & 0.951 & 0.964 \\ 
        4 & 0.976 & 0.948 & 0.972 & 0.984 & 0.968 & 0.959 & 0.968 \\ 
        5 & 0.975 & 0.944 & 0.950 & 0.981 & 0.976 & 0.947 & 0.962 \\ 
        6 & 0.973 & 0.919 & 0.950 & 0.978 & 0.975 & 0.952 & 0.958 \\ 
        7 & 0.979 & 0.911 & 0.968 & 0.974 & 0.958 & 0.952 & 0.957 \\ 
        8 & 0.981 & 0.892 & 0.953 & 0.977 & 0.973 & 0.959 & 0.956 \\ 
        9 & 0.983 & 0.875 & 0.967 & 0.974 & 0.980 & 0.959 & 0.956 \\ 
        10 & 0.980 & 0.857 & 0.947 & 0.967 & 0.957 & 0.958 & 0.944 \\ 
        11 & 0.984 & 0.847 & 0.931 & 0.944 & 0.964 & 0.959 & 0.938 \\ 
        12 & 0.990 & 0.828 & 0.865 & 0.920 & 0.974 & 0.959 & 0.923 \\ 
        13 & 0.990 & 0.953 & 0.901 & 0.865 & 0.968 & 0.959 & 0.939 \\ 
        14 & 0.990 & 0.933 & 0.935 & 0.874 & 0.975 & 0.957 & 0.944 \\ 
        15 & 0.988 & 0.919 & 0.945 & 0.858 & 0.972 & 0.959 & 0.940 \\ 
        16 & 0.977 & 0.900 & 0.941 & 0.854 & 0.966 & 0.957 & 0.933 \\ 
        17 & 0.974 & 0.899 & 0.944 & 0.883 & 0.948 & 0.955 & 0.934 \\ 
        18 & 0.978 & 0.897 & 0.946 & 0.892 & 0.930 & 0.957 & 0.933 \\ 
        19 & 0.951 & 0.882 & 0.938 & 0.874 & 0.913 & 0.957 & 0.919 \\ 
        20 & 0.947 & 0.885 & 0.900 & 0.857 & 0.858 & 0.956 & 0.900 \\ 
        21 & 0.932 & 0.879 & 0.887 & 0.808 & 0.740 & 0.957 & 0.867 \\ 
        22 & 0.927 & 0.858 & 0.927 & 0.789 & 0.668 & 0.957 & 0.854 \\ 
        23 & 0.859 & 0.827 & 0.889 & 0.862 & 0.579 & 0.957 & 0.829 \\ 
        24 & 0.872 & 0.825 & 0.867 & 0.808 & 0.502 & 0.954 & 0.805 \\ \hline 
    \end{tabular}
    }
     \caption{Distance Effect: Averaged (across inputs) $R^2$ values of different \emph{Larger variants of} LLMs for different layers when fitting a linear function. RoB: Roberta-base model, BERT: uncased variant.} 
\label{tab:distance_effect_model_layers_large}
\end{table}

\renewcommand*{\MinNumber}{0.505}%
\renewcommand*{\MaxNumber}{0.65}%

\begin{table}[!h]
    \centering
    \resizebox{0.48\textwidth}{!}{%
    \begin{tabular}{cccccccR}
    \hline
        \multicolumn{1}{c}{\begin{tabular}[c]{@{}c@{}} Layer\end{tabular}} & T5 & BART & RoB & XLNET & \multicolumn{1}{c}{\begin{tabular}[c]{@{}c@{}} BERT\end{tabular}} & GPT-2 & \multicolumn{1}{c}{\begin{tabular}[C]{@{}c@{}} Avg.\end{tabular}} \\ \hline
          1 & 0.785 & 0.800 & 0.591 & 0.630 & 0.608 & 0.467 & 0.647 \\ 
        2 & 0.794 & 0.763 & 0.275 & 0.666 & 0.198 & 0.597 & 0.549 \\ 
        3 & 0.894 & 0.709 & 0.379 & 0.665 & 0.214 & 0.661 & 0.587 \\ 
        4 & 0.922 & 0.719 & 0.465 & 0.661 & 0.345 & 0.620 & 0.622 \\ 
        5 & 0.940 & 0.721 & 0.550 & 0.634 & 0.387 & 0.563 & 0.632 \\ 
        6 & 0.925 & 0.606 & 0.426 & 0.644 & 0.661 & 0.584 & 0.641 \\ 
        7 & 0.912 & 0.441 & 0.360 & 0.603 & 0.636 & 0.594 & 0.591 \\ 
        8 & 0.923 & 0.399 & 0.460 & 0.548 & 0.726 & 0.595 & 0.608 \\ 
        9 & 0.915 & 0.354 & 0.435 & 0.541 & 0.750 & 0.599 & 0.599 \\ 
        10 & 0.923 & 0.329 & 0.546 & 0.553 & 0.727 & 0.593 & 0.612 \\ 
        11 & 0.924 & 0.362 & 0.458 & 0.574 & 0.727 & 0.601 & 0.608 \\ 
        12 & 0.890 & 0.351 & 0.512 & 0.543 & 0.728 & 0.601 & 0.604 \\ 
        13 & 0.864 & 0.801 & 0.467 & 0.468 & 0.757 & 0.595 & 0.659 \\ 
        14 & 0.837 & 0.861 & 0.452 & 0.436 & 0.751 & 0.600 & 0.656 \\ 
        15 & 0.805 & 0.796 & 0.480 & 0.454 & 0.741 & 0.597 & 0.645 \\ 
        16 & 0.761 & 0.683 & 0.449 & 0.436 & 0.739 & 0.597 & 0.611 \\ 
        17 & 0.692 & 0.550 & 0.391 & 0.423 & 0.746 & 0.598 & 0.567 \\ 
        18 & 0.743 & 0.520 & 0.453 & 0.423 & 0.747 & 0.594 & 0.580 \\ 
        19 & 0.633 & 0.512 & 0.435 & 0.391 & 0.788 & 0.594 & 0.559 \\ 
        20 & 0.583 & 0.513 & 0.448 & 0.373 & 0.828 & 0.596 & 0.557 \\ 
        21 & 0.523 & 0.532 & 0.512 & 0.345 & 0.847 & 0.592 & 0.558 \\ 
        22 & 0.432 & 0.546 & 0.633 & 0.350 & 0.874 & 0.588 & 0.571 \\ 
        23 & 0.356 & 0.455 & 0.491 & 0.316 & 0.846 & 0.590 & 0.509 \\ 
        24 & 0.335 & 0.444 & 0.634 & 0.250 & 0.801 & 0.584 & 0.508 \\ \hline 
    \end{tabular}
    }
     \caption{Size Effect: Averaged (across inputs) $R^2$ values of different \emph{Larger variants of} LLMs for different layers when fitting a linear function. RoB: Roberta-base model, BERT: uncased variant.} 
\label{tab:size_effect_model_layers_large}
\end{table}

\begin{table}[!ht]
    \centering
        \resizebox{0.41\textwidth}{!}{%

    \begin{tabular}{lcccc}
    \hline
    \multicolumn{1}{l}{\begin{tabular}[l]{@{}l@{}} LLMs$\backslash$Input\end{tabular}} 
         & \multicolumn{1}{c}{\begin{tabular}[c]{@{}c@{}} LC \end{tabular}}  & \multicolumn{1}{c}{\begin{tabular}[c]{@{}c@{}} MC\end{tabular}}  & \multicolumn{1}{c}{\begin{tabular}[c]{@{}c@{}} Digits\end{tabular}}  & \multicolumn{1}{c}{\begin{tabular}[c]{@{}c@{}} Avg. \end{tabular}} \\ \hline
        T5 & 0.868 & 0.816 & 0.833 & 0.839 \\ 
        BART & 0.767 & 0.838 & 0.478 & 0.694 \\ 
        RoBERTa & 0.672 & 0.686 & 0.725 & 0.694 \\ 
        XLNET & 0.617 & 0.649 & 0.711 & 0.659 \\ 
        BERT (uncased) & \multicolumn{2}{c}{0.786} & 0.732 & 0.759 \\ 
        GPT2 & 0.669 & 0.720 & 0.767 & 0.718 \\ \hline

        \multicolumn{1}{l}{\begin{tabular}[l]{@{}l@{}} Total Averages \\across models\end{tabular}}
         & 0.730 & 0.749 & 0.707 & 0.718 \\ \hline
    \end{tabular}
    }
    \caption{Ratio Effect: Averaged (across layers) $R^2$ values of different \emph{Larger variants of} LLMs on different input types when fitting a negative exponential function. LC: Lowercase number words, MC: Mixedcase number words.}
\label{tab:ratio_effect_models_large}
\end{table}

\renewcommand*{\MinNumber}{0.620}%
\renewcommand*{\MaxNumber}{0.826}%

\begin{table}[!h]
    \centering
    \resizebox{0.48\textwidth}{!}{%
    \begin{tabular}{cccccccR}
    \hline
        \multicolumn{1}{c}{\begin{tabular}[c]{@{}c@{}} Layer\end{tabular}} & T5 & BART & RoB & XLNET & \multicolumn{1}{c}{\begin{tabular}[c]{@{}c@{}} BERT \end{tabular}} & GPT-2 & \multicolumn{1}{c}{\begin{tabular}[C]{@{}c@{}} Avg.\end{tabular}} \\ \hline
           1 & 0.868 & 0.837 & 0.803 & 0.881 & 0.829 & 0.733 & 0.825 \\ 
        2 & 0.803 & 0.740 & 0.529 & 0.873 & 0.657 & 0.708 & 0.718 \\ 
        3 & 0.792 & 0.798 & 0.573 & 0.875 & 0.602 & 0.775 & 0.736 \\ 
        4 & 0.828 & 0.782 & 0.722 & 0.868 & 0.667 & 0.725 & 0.765 \\ 
        5 & 0.860 & 0.823 & 0.716 & 0.863 & 0.664 & 0.652 & 0.763 \\ 
        6 & 0.878 & 0.811 & 0.671 & 0.836 & 0.765 & 0.680 & 0.774 \\ 
        7 & 0.898 & 0.686 & 0.669 & 0.818 & 0.735 & 0.704 & 0.752 \\ 
        8 & 0.896 & 0.657 & 0.726 & 0.797 & 0.722 & 0.716 & 0.753 \\ 
        9 & 0.910 & 0.658 & 0.714 & 0.792 & 0.838 & 0.729 & 0.773 \\ 
        10 & 0.915 & 0.639 & 0.718 & 0.774 & 0.818 & 0.729 & 0.766 \\ 
        11 & 0.921 & 0.640 & 0.583 & 0.745 & 0.835 & 0.725 & 0.742 \\ 
        12 & 0.917 & 0.638 & 0.518 & 0.691 & 0.868 & 0.724 & 0.726 \\ 
        13 & 0.920 & 0.836 & 0.538 & 0.593 & 0.820 & 0.728 & 0.739 \\ 
        14 & 0.937 & 0.764 & 0.679 & 0.585 & 0.837 & 0.724 & 0.755 \\ 
        15 & 0.931 & 0.715 & 0.772 & 0.546 & 0.822 & 0.722 & 0.751 \\ 
        16 & 0.915 & 0.713 & 0.762 & 0.514 & 0.815 & 0.726 & 0.741 \\ 
        17 & 0.904 & 0.684 & 0.747 & 0.492 & 0.836 & 0.718 & 0.730 \\ 
        18 & 0.907 & 0.666 & 0.728 & 0.497 & 0.815 & 0.728 & 0.723 \\ 
        19 & 0.778 & 0.617 & 0.754 & 0.464 & 0.807 & 0.720 & 0.690 \\ 
        20 & 0.754 & 0.613 & 0.717 & 0.450 & 0.775 & 0.720 & 0.671 \\ 
        21 & 0.692 & 0.600 & 0.723 & 0.435 & 0.699 & 0.716 & 0.644 \\ 
        22 & 0.679 & 0.605 & 0.802 & 0.459 & 0.715 & 0.721 & 0.664 \\ 
        23 & 0.637 & 0.587 & 0.730 & 0.478 & 0.651 & 0.716 & 0.633 \\ 
        24 & 0.592 & 0.559 & 0.767 & 0.486 & 0.624 & 0.703 & 0.622 \\ \hline 
    \end{tabular}
    }
     \caption{Ratio Effect: Averaged (across inputs) $R^2$ values of different \emph{Larger variants of} LLMs for different layers when fitting a negative exponential function. RoB: Roberta-base model, BERT: uncased variant.} 
\label{tab:ratio_effect_model_layers_large}
\end{table}

\begin{table}[!ht]
    \centering
        \resizebox{0.41\textwidth}{!}{%

    \begin{tabular}{lcccc}
    \hline
    \multicolumn{1}{l}{\begin{tabular}[l]{@{}l@{}} LLMs$\backslash$Input\end{tabular}} 
         & \multicolumn{1}{c}{\begin{tabular}[c]{@{}c@{}} LC \end{tabular}}  & \multicolumn{1}{c}{\begin{tabular}[c]{@{}c@{}} MC\end{tabular}}  & \multicolumn{1}{c}{\begin{tabular}[c]{@{}c@{}} Digits\end{tabular}}  & \multicolumn{1}{c}{\begin{tabular}[c]{@{}c@{}} Avg. \end{tabular}} \\ \hline
        T5 & 0.572 & 0.127 & 0.408 & 0.369 \\ 
        BART & 0.677 & 0.546 & 0.515 & 0.580 \\ 
        RoBERTa & 0.669 & 0.573 & 0.473 & 0.572 \\ 
        XLNET & 0.498 & 0.373 & 0.465 & 0.445 \\ 
        BERT (uncased) & \multicolumn{2}{c}{0.519} & 0.541 & 0.530 \\ 
        GPT2 & 0.623 & 0.624 & 0.888 & 0.711 \\ \hline

        \multicolumn{1}{l}{\begin{tabular}[l]{@{}l@{}} Total Averages \\across models\end{tabular}}
         & 0.593 & 0.460 & 0.548 & 0.534 \\ \hline
    \end{tabular}
    }
    \caption{Averaged (across layers) correlation values when comparing MDS values with $Log_{10}1$ to $Log_{10}9$ for \emph{Large variants of} different LLMs. LC: Lowercase number words, MC: Mixedcase number words.}
\label{tab:mds_effect_models_large}
\end{table}

\renewcommand*{\MinNumber}{0.410}%
\renewcommand*{\MaxNumber}{0.645}%
\begin{table}[!h]
    \centering
    \resizebox{0.48\textwidth}{!}{%
    \begin{tabular}{cccccccR}
    \hline
        \multicolumn{1}{c}{\begin{tabular}[c]{@{}c@{}} Layer\end{tabular}} & T5 & BART & RoB & XLNET & \multicolumn{1}{c}{\begin{tabular}[c]{@{}c@{}} BERT\end{tabular}} & GPT-2 & \multicolumn{1}{l}{\begin{tabular}[C]{@{}c@{}}Avg. \end{tabular}} \\ \hline
           1 & 0.675 & 0.633 & 0.731 & 0.590 & 0.542 & 0.689 & 0.643 \\ 
        2 & 0.249 & 0.662 & 0.461 & 0.649 & 0.555 & 0.767 & 0.557 \\ 
        3 & 0.251 & 0.673 & 0.522 & 0.689 & 0.662 & 0.707 & 0.584 \\ 
        4 & 0.156 & 0.682 & 0.698 & 0.674 & 0.353 & 0.703 & 0.544 \\ 
        5 & 0.059 & 0.518 & 0.493 & 0.686 & 0.065 & 0.719 & 0.423 \\ 
        6 & 0.219 & 0.471 & 0.411 & 0.533 & 0.535 & 0.729 & 0.483 \\ 
        7 & 0.569 & 0.421 & 0.558 & 0.549 & 0.367 & 0.688 & 0.526 \\ 
        8 & 0.578 & 0.413 & 0.540 & 0.690 & 0.385 & 0.695 & 0.550 \\ 
        9 & 0.581 & 0.710 & 0.594 & 0.546 & 0.598 & 0.720 & 0.625 \\ 
        10 & 0.495 & 0.716 & 0.531 & 0.487 & 0.710 & 0.718 & 0.610 \\ 
        11 & 0.286 & 0.691 & 0.404 & 0.495 & 0.576 & 0.702 & 0.526 \\ 
        12 & 0.481 & 0.682 & 0.304 & 0.466 & 0.708 & 0.700 & 0.557 \\ 
        13 & 0.387 & 0.605 & 0.533 & 0.394 & 0.588 & 0.721 & 0.538 \\ 
        14 & 0.483 & 0.672 & 0.538 & 0.383 & 0.574 & 0.718 & 0.562 \\ 
        15 & 0.486 & 0.386 & 0.596 & 0.241 & 0.586 & 0.705 & 0.500 \\ 
        16 & 0.485 & 0.454 & 0.689 & 0.140 & 0.591 & 0.692 & 0.509 \\ 
        17 & 0.536 & 0.677 & 0.617 & 0.163 & 0.588 & 0.719 & 0.550 \\ 
        18 & 0.259 & 0.562 & 0.651 & 0.251 & 0.602 & 0.704 & 0.505 \\ 
        19 & 0.458 & 0.750 & 0.583 & 0.077 & 0.599 & 0.694 & 0.527 \\ 
        20 & 0.463 & 0.545 & 0.652 & 0.246 & 0.585 & 0.718 & 0.535 \\ 
        21 & 0.362 & 0.526 & 0.653 & 0.524 & 0.554 & 0.700 & 0.553 \\ 
        22 & 0.402 & 0.522 & 0.656 & 0.247 & 0.596 & 0.719 & 0.524 \\ 
        23 & -0.019 & 0.466 & 0.649 & 0.490 & 0.600 & 0.720 & 0.484 \\ 
        24 & -0.051 & 0.473 & 0.652 & 0.476 & 0.205 & 0.726 & 0.414 \\ \hline 
    \end{tabular}
    }
     \caption{Averaged (across inputs) correlation values of the \emph{Large variants} of different LLMs on different model layers when comparing MDS values with $Log_{10}1$ to $Log_{10}9$. RoB: Roberta-base model, BERT: uncased variant.} 
\label{tab:mds_effect_model_layers_large}
\end{table}
For the models in Table\ref{tab:models}, we show the three effects for the larger variants. The variants have the same architectures and training methodologies as their base variants but more parameters (~thrice the number of parameters). The in-depth results for the three effects are presented in tables \ref{tab:distance_effect_models_large}, \ref{tab:distance_effect_model_layers_large}, \ref{tab:size_effect_models_large}, \ref{tab:size_effect_model_layers_large}, \ref{tab:ratio_effect_models_large}, and \ref{tab:ratio_effect_model_layers_large}. We also present the MDS correlation values in the same manner as done for base variants; see tables \ref{tab:mds_effect_models_large} and \ref{tab:mds_effect_model_layers_large}.

Given the large layer count for these model variants and the multiple tables, we also present a summarized view of the results in tables \ref{tab:inputs_large}, \ref{tab:mds_residual_large}, \ref{tab:layers_large}.
\subsection{Cased vs Uncased BERT}
\label{case_uncased}

\begin{table}[!h]
    \centering
     \resizebox{0.46\textwidth}{!}{%
    \begin{tabular}{llcccc}
    \hline
        Variant &  Effect & \multicolumn{1}{c}{\begin{tabular}[c]{@{}c@{}} LC\end{tabular}}  & \multicolumn{1}{c}{\begin{tabular}[c]{@{}c@{}} MC\end{tabular}}  & \multicolumn{1}{c}{\begin{tabular}[c]{@{}c@{}} Digits\end{tabular}}  & \multicolumn{1}{c}{\begin{tabular}[c]{@{}c@{}} Avg. \end{tabular}} \\ \hline
        \multirow{4}{*}{Uncased} & Distance & \multicolumn{2}{c}{0.976} & 0.944 & 0.960  \\ 
         & Size & \multicolumn{2}{c}{0.803} & 0.851 & 0.827  \\ 
        & Ratio  & \multicolumn{2}{c}{0.906} & 0.757 & 0.831  \\ 
        & MDS (Corr.) & \multicolumn{2}{c}{0.312} & 0.423 & 0.386  \\ \hline
        \multirow{4}{*}{Cased} & Distance & 0.958 & 0.980 & 0.890 & 0.943  \\ 
         & Size  & 0.664 & 0.691 & 0.918 & 0.758 \\ 
        & Ratio  & 0.854 & 0.880 & 0.866 & 0.867 \\ 
        & MDS (Corr.) & 0.621 & 0.553 & 0.487 & 0.554\\  \hline
    \end{tabular}
    }
    \caption{Behavioral differences between the cased and uncased variants of the BERT architecture. LC: Lowercase number words, MC: Mixed-case number words.}
    \label{tab:cased_uncased}
\end{table}
The behavioral differences between the cased and uncased variants of the BERT architecture are shown in Table\ref{case_uncased}. Despite different preprocessing paradigms, both models have similar performance characteristics. The only visible distinction is the higher correlation values for the cased version when compared to the uncased version of the model. 
\subsection{Impact of Distance effect and Size effect in Ratio effect scores}
\begin{table}[!h]
    \centering
     \resizebox{0.46\textwidth}{!}{%
    \begin{tabular}{llcccc}
    \hline
        Variant & & Coef. & Std. Error & t Stat & P-value \\ \hline
        & Intercept & -0.916 & 0.531 & -1.722 & 0.119 \\ 
        Base & Distance Effect  & 1.953 & 0.452 & 4.314 & \textbf{0.001} $\odot$ \\ 
        & Size Effect  & -0.228 & 0.188 & -1.212 & 0.256 \\ \hline
        & Intercept & -0.188 & 0.075 & -2.491 & 0.0.021 \\ 
        Large & Distance Effect  & 0.700 & 0.117 & 5.997 & \textbf{0.000} $\oplus$ \\ 
        & Size Effect  & 0.447 & 0.124 & 3.612 & \textbf{0.001} $\odot$ \\ \hline
    \end{tabular}
    }
    \caption{Impact of layer-wise trends of distance and size effect on the ratio effect; $\odot$ indicates statistical significance with p-value less that 0.01, $\oplus$ indicates statistical significance with p-value less that 0.00001 }
    \label{tab:regression_ratio}
\end{table}

When interpreting LLM findings on the ratio effect, we observe that they are dominated by the distance effect as compared to the size effect. We observe the same decreasing trend in averaged results over input types in layers; see Table \ref{tab:ratio_effect_model_layers} (column: Total Averages). The impact of layer-wise trends can be quantified using regression with the distance effect and size effect as inputs (column: Total Averages; tables \ref{tab:distance_effect_model_layers}, \ref{tab:size_effect_model_layers}) and the ratio effect (column: Total Averages; Table\ref{tab:size_effect_model_layers}) as output. Importantly, the distance effect averages are statistically significant predictors of ratio effect averages; see Table \ref{tab:regression_ratio}). These results provide a superficial view of the impact of distance and size effect in the ratio effect scores because of the aggregation performed at different levels of the study.

\end{document}